\pdfoutput=1
\documentclass[11pt]{article}

\usepackage[]{StyleFiles/EMNLP2023}

\usepackage{times}
\usepackage{latexsym}

\usepackage{amsmath}
\usepackage{caption}
\usepackage{subcaption}
\usepackage{hyperref}
\usepackage{url}
\usepackage{lipsum}
\usepackage{graphicx}
\usepackage{tabularx}
\usepackage{array}
\usepackage{multirow}
\usepackage{tikz}
\usepackage{pgf-pie}

\usepackage{graphicx}
\usepackage{tabularx}
\usepackage{soul}
\usepackage{caption}
\usepackage{subcaption}
\usepackage{latexsym}
\usepackage{pgfplots}
\usepackage{amsmath}
\usepackage{caption}
\usepackage{subcaption}
\usepackage{hyperref}
\usepackage{url}
\usepackage{lipsum}
\usepackage{graphicx}
\usepackage{tabularx}
\usepackage{array}
\usepackage{multirow}
\usepackage{booktabs}

\usepackage[T1]{fontenc}
\usepackage[utf8]{inputenc}

\usepackage{microtype}

\usepackage{inconsolata}

\title{\textsc{Fin-Fact}: A Benchmark Dataset for Multimodal Financial Fact Checking and Explanation Generation}

\author{\textbf{Aman Rangapur$^1$} \quad \textbf{Haoran Wang$^1$} \quad \textbf{Ling Jian$^2$} \quad \textbf{Kai Shu$^1$} \\
        $^1$Illinois Institute of Technology, Chicago, IL, USA \\
        $^2$School of Economics and Management, China University of Petroleum, Qingdao, China\\
        \texttt{\{arangapur, hwang219\}@hawk.iit.edu}, \texttt{bebetter@upc.edu.cn}, 
        \texttt{kshu@iit.edu}}
\begin{document}
\maketitle
\begin{abstract}
Misinformation can have detrimental effects on various aspects of our society. 
The recent incident, where a cryptocurrency exchange company failed, has led to substantial losses for investors. There is compelling evidence indicating that the widespread dissemination of financial misinformation played a crucial role in this situation.
Recognizing the lack of benchmark datasets specifically designed to tackle financial misinformation, we have developed \textsc{Fin-Fact}, a comprehensive benchmark dataset curated for multimodal fact-checking and explanation generation. 
What sets \textsc{Fin-Fact} apart is its inclusion of expert fact-checker annotations and detailed justifications, infusing the dataset with both expertise and credibility. \textsc{Fin-Fact} adopts a multimodal approach, encompassing not only textual but also visual content, thereby providing complementary information sources for a more holistic analysis of factuality. Beyond this, the dataset offers insightful explanations accompanying each fact-check, empowering users to delve into the rationale behind fact-checking decisions and it can be used for automated systems to generate explanations. This transparency not only validates the credibility of the claims but also fosters trust in the overall fact-checking process. The \textsc{Fin-Fact} dataset, along with our experimental codes is available at \url{https://github.com/IIT-DM/Fin-Fact/}.
\end{abstract}

\section{Introduction}
In an era characterized by the rapid spread of misinformation and the proliferation of fake news, fact-checking has emerged as a critical tool for ensuring the accuracy and reliability of information \cite{saakyan-etal-2021-covid,wadden-etal-2020-fact,sarrouti-etal-2021-evidence-based}. The emergence of social media platforms and the wide accessibility of multimodal content have intensified the complexities linked with verifying the accuracy of assertions \cite{mishra2022factify}. Notably, the financial sector introduces its distinctive array of difficulties, given that precise and timely data plays a pivotal role in enabling well-informed investment choices and upholding market stability. Additionally, financial fact-checking encounters specific challenges, such as the need for customized data to address unique requirements and nuances. 
Furthermore, the manipulation of images to exploit visualization bias presents another significant challenge in the verification process \cite{mansoor2018data,husser2014investors}.

\begin{figure}
    \centering
    \includegraphics[width=1\linewidth]{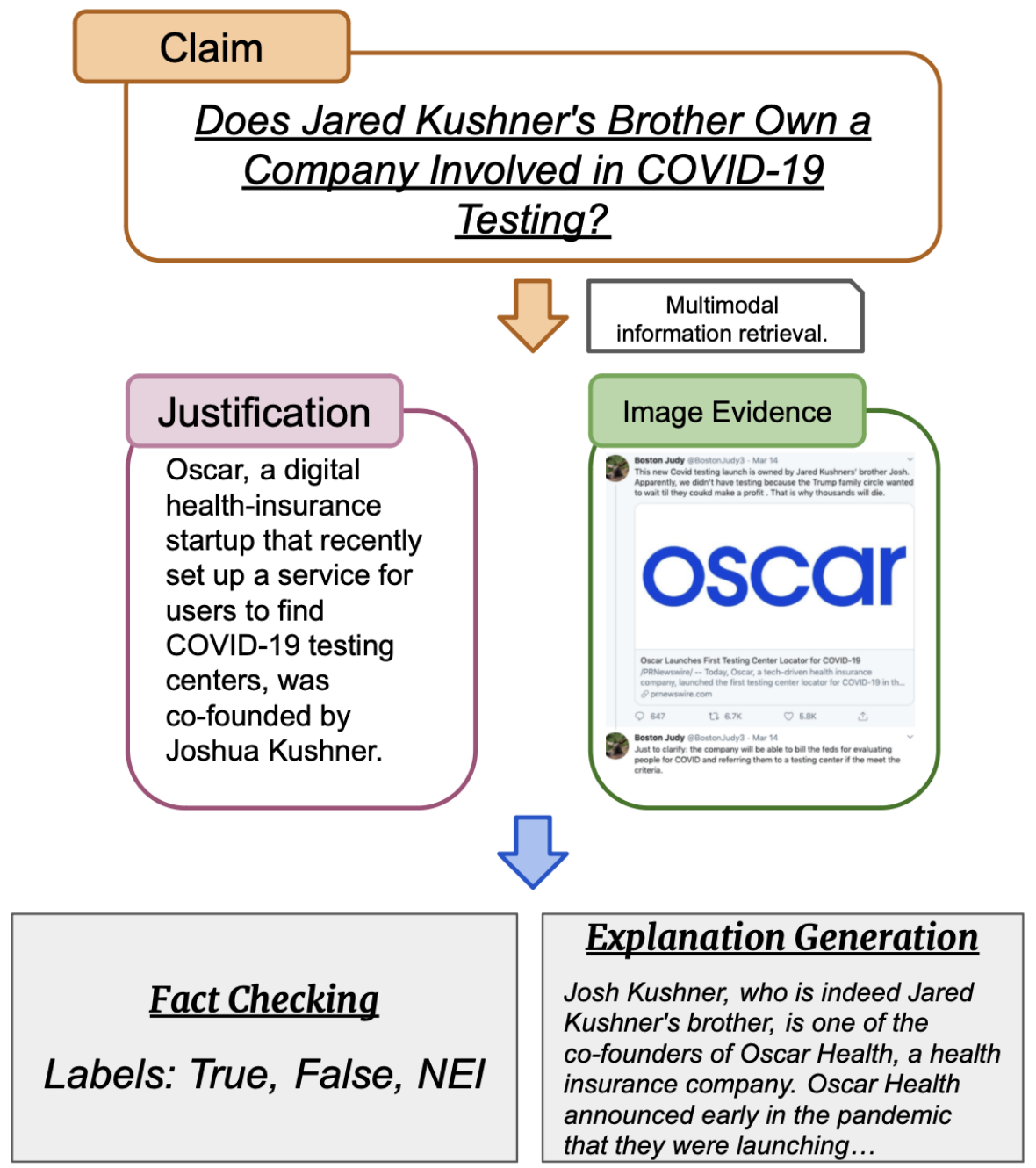}
    \caption{Illustration of comprehensive multimodal fact-checking, including True, False, and Not Enough Information (NEI), alongside the creation of explanations.}
    \label{fig:example}
\end{figure}

The rise of misinformation in the financial domain has become a pressing concern, with potential impacts on public trust, investor decisions, and overall market stability \cite{kogan2019fake,clarke2018fake,zhi2021financial,liu2022role02}. To counter the spread of misleading information, fact-checking methods have gained importance in financial reporting and analysis \cite{zhi2021financial, mohankumar2023financial}. However, the development of reliable models in this domain has been hindered by the lack of suitable benchmark datasets that accurately represent the intricacies of financial information and context \cite{rangapur2023investigating}.

\begin{table}[!htbp]
\centering
\caption{An example instance from \textsc{Fin-Fact} dataset.}
\resizebox{1.0\linewidth}{!}{%
\begin{tabular}{m{5em} m{18em}}
\hline
\textbf{Feature} & \textbf{Example}\\
\hline
\textbf{Claim} & Did Biden Call for an `End to Shareholder Capitalism'? \\
\textbf{Author} & Dan MacGuill \\
\textbf{Posted} & 08/06/2020 \\
\textbf{Sci-Digest} & Fear-mongering Facebook memes misrepresented the position articulated by the presumptive Democratic presidential nominee in a speech in July 2020.\\
\textbf{Justification} & In the summer of 2020, multiple readers asked Snopes to investigate claims that former...\\
\textbf{Evidence} & In July and August, Facebook users shared posts that contained the following text...\\
\textbf{Image} & https://drive.google.com/file/1UTniyHzAB8 \\
\textbf{Issues} & Debt \\
\textbf{Label} & False\\
\hline
\end{tabular}
}
\label{table:labels_finfact}
\end{table}

In recent years, there has been notable progress in creating various datasets for fact-checking \cite{wadden-etal-2022-scifact, sarrouti-etal-2021-evidence-based, saakyan-etal-2021-covid}. However, there is a noticeable gap in addressing the unique demands of fact-checking within the financial domain. Financial fact-checking faces several significant challenges. \textbf{Firstly}, it requires meticulously curated data that can encompass the intricate nuances of financial discourse. Financial documents and journalistic pieces often employ specialized language that differs from conventional structures. However, existing datasets frequently lack comprehensive coverage of financial news articles, and the absence of expert annotations diminishes the reliability of the data. \textbf{Secondly}, financial data is highly context-sensitive and constantly evolving, emphasizing the need for a dataset that can accurately capture the dynamic nature of financial markets. \textbf{Lastly}, the landscape of financial fact-checking introduces the challenge of visualization, where deliberate manipulation of visual content can shape perception and distort the accuracy of claims.

In this paper, we tackle the challenge of compiling, annotating, and refining a comprehensive corpus of financial texts that faithfully represents financial reporting, accounting methodologies, and market fluctuations. The realms of financial fact-checking and explanation generation present distinct obstacles that require specialized approaches. The necessity for tailored data capable of navigating financial terminology and intricate visual elements underscores the interdisciplinary nature inherent in this research endeavor. Figure \ref{fig:example} illustrates a comprehensive multimodal fact-checking and the creation of explanations while Table \ref{table:labels_finfact} displays an example instance from the corpus.

We introduce \textsc{Fin-Fact}, a new benchmark dataset created specifically for the task of multimodal financial fact-checking and explanation generation. Our contributions are as follows:
\vspace{-0.2cm}
\begin{itemize}
    \item We introduce \textsc{Fin-Fact}, the benchmark dataset designed to verify multimodal claims within the financial domain.
    \item \textsc{Fin-Fact} enables explanation generation through expert comments provided by fact-checking professionals.
    \item Our investigation show that the state-of-the-art models have difficulty performing well on \textsc{Fin-Fact} in the open-domain context, indicating the need to improve how these systems handle different types of data.
\end{itemize}

\begin{table*}[!htbp]
\caption{Existing Datasets on Fact Checking.}
\resizebox{\textwidth}{!}{%
\begin{tabular}{@{}llllll@{}}
\toprule
Dataset & Topic & Size(\# of instances) & Modality & Source & \# of classes \\ \midrule
LIAR  & Political & 12,836 & Text & Politifact & 6 \\ \midrule
FEVER & Multi-Domain & 185,445 & Text & Wikipedia & 3 \\ \midrule
FakeNewsNet  & Political \& Celebrity  & 602,659  & Text & Twitter  & 2 \\ \midrule
PHEME  & Multi-Domain & 330  & Text & Twitter   & 2 \\ \midrule
MM-COVID    & Medical & 11,173 & Text, Image & \begin{tabular}[c]{@{}l@{}}Politifact, Snopes, \\ Poynter \& Fullfact\end{tabular} & 2 \\ \midrule
Factify  & Multi-Domain       & 50,000  & Text, Image  & Twitter  & 5 \\ \midrule
Fauxtography  & Multi-Domain       & 1,223  & Text, Image  & Snopes, Reuters  & 2 \\ \midrule
MOCHEG  & Multi-Domain & 15,601  & Text, Image & Politifact, Snopes  & 3 \\ \midrule
Fakeddit  & Multi-Domain  & 1,063,106  & Text, Image & Reddit & 2,3,6 \\ \midrule
\textsc{Fin-Fact} (ours) & Finance   & 3,369  & Text, Image  & \begin{tabular}[c]{@{}l@{}}Politifact, Snopes\\  \& FactCheck\end{tabular}  & 3 \\
\bottomrule

\end{tabular}
}
\label{factcheck_datasets_list}
\end{table*}
\section{Related Work}
\textbf{Fact Checking.} Significant efforts have been dedicated to creating fact-checking datasets for automated fact-checking systems \cite{wadden-etal-2020-fact, wadden-etal-2022-scifact, thorne-etal-2018-fever, saakyan-etal-2021-covid}. Previous studies have predominantly focused on predicting the accuracy of claims from diverse sources. While large-scale datasets from various domains have been utilized \cite{gupta2021xfact}, they might not be suitable for identifying misinformation related to financial matters due to domain-specific disparities. A few existing datasets on fact checking are listed in Table \ref{factcheck_datasets_list}.

Although general-content misinformation datasets are readily accessible, only a limited number of datasets pertain to online financial misinformation \cite{clarke2018fake, hossain2020banfakenews, kogan2019fake, zhi2021financial, liu2022role02, zhang2022theory009, NBERw30994}. Current financial misinformation datasets lack clear labeling and justifications, raising concerns about result reliability. In contrast, the \textsc{Fin-Fact} dataset is distinct with genuine data and a multimodal structure, combining text and images to encompass a wide range of financial information. Additionally, it includes expert fact-checker comments, enabling comprehensive explanations by models.\\

\textbf{Explanation Generation.} Explanation generation plays a pivotal role in facilitating human comprehension of claim credibility. It involves leveraging external knowledge graphs to create semantic traces originating from the claim itself \cite{gad2019exfakt,li2020mm,sarrouti-etal-2021-evidence-based}. These semantic traces function as explanations that substantiate the veracity of claims. This approach offers valuable insights into the rationale behind the model's decision-making, thereby fostering trust. Moreover, the process of explanation generation relies on drawing evidence from diverse sources \cite{atanasova-etal-2020-generating-fact, hanselowski2019richly, fan2020generating} to validate claims. However, this evidence is frequently comprised of isolated sentences extracted from extensive collections of documents, which can make it challenging for humans to interpret the broader context. To effectively generate explanations, a high-quality dataset annotated by humans is essential. This paper addresses the need for such a dataset.
\section{The \textsc{Fin-Fact} Dataset}
% \vspace{-0.4cm}
The \textsc{Fin-Fact} dataset presents a diverse array of labels that enhance the depth of analysis when evaluating financial claims. These labels contribute a multifaceted perspective to the fact-checking process, augmenting its analytical capabilities.

The dataset comprises essential attributes, including the \textit{Claim} and \textit{Author} labels, which respectively represent the core assertion and its source. Temporal context is introduced via the \textit{Posted Date} attribute, while claim summaries are provided by the \textit{Sci-digest} label. To further contextualize claims, \textit{Justification} offers insights into their accuracy, \textit{Evidence} presents supporting information linked through \textit{Evidence link}, and \textit{Image link} to address the visual dimension, and the \textit{Issues} label highlights complexities within claims. Ultimately, the \textit{Claim Label} categorizes claims as \texttt{True}, \texttt{False}, or \texttt{NEI (Not Enough Information)}.

By amalgamating these labels, the dataset establishes a comprehensive and multidimensional resource. This resource accommodates textual, temporal, evidentiary, and visual components, all of which are imperative for a thorough evaluation of claims during the fact-checking process.

\subsection{Data Collection and Preprocessing}
PolitiFact\footnote{\url{http://politifact.com/}}, Snopes\footnote{\url{http://snopes.com/}} and FactCheck\footnote{\url{http://factcheck.org}} are prominent online platforms dedicated to countering the spread of false information. These platforms engage professional fact-checkers to meticulously analyze and verify individual claims, subsequently producing articles that offer their conclusions supported by relevant evidence. In our study, we leveraged these platforms as our primary sources of data.

To elaborate, we devised a systematic process to gather essential information from PolitiFact, Snopes and FactCheck websites. This encompassed the extraction of text-based claims and the assignment of corresponding truthfulness labels. Moreover, we retrieved both textual and visual evidence, along with their associated links, which contributed substantially to the assessment of claim accuracy. Through advanced keyword-based filtering mechanisms, we identified and isolated claims and articles that specifically pertained to financial domain. This process involved the careful selection of terms and phrases related to various financial sectors, including investment, taxation, and corporate finance. Furthermore, we ensured that the extracted claims were diverse in nature, encompassing a wide array of financial topics such as market trends, economic policies, financial disclosures, and fiscal regulations. By emphasizing this specialized filtration approach, we aimed to create a dataset that distinctly focuses on the intricacies of fact-checking within the financial landscape.

It's noteworthy that the initial claims were collected by journalists affiliated with these platforms. These claims originated from diverse sources, including online speeches, public statements, news articles, and social media posts. Importantly, the fact-checkers from these platforms played a pivotal role by providing truthfulness labels, pertinent evidence, references to corroborating sources, and the articles delivering their final verdict. This comprehensive approach ensured the thorough and reliable collection of data, reinforcing the credibility of our assessment of claim accuracy.

Following the data collection, we embarked on a rigorous data cleaning process to ensure the quality and reliability of our dataset. This process was crucial as it eliminated potential noise or inconsistencies in the data that could affect subsequent analysis. We removed any extra spaces within and between the claims to standardize the text data and facilitate easier text processing in later stages. Emojis were eliminated from the text. While emojis can sometimes convey meaningful information, their interpretation can be highly subjective and context-dependent. To maintain the objectivity of our analysis, we decided to exclude them from our dataset. Double quotations were removed from the text. These can often create confusion during text processing, especially when they are used within sentences. By removing them, we ensured that each claim is treated as a single continuous string of text, thereby simplifying the subsequent text analysis. 
Additional cleaning steps tailored to the specific characteristics of our dataset were conducted. These included the removal of HTML tags, correction of misspelled words, and standardization of text casing. Through these meticulous data cleaning procedures, we were able to refine our dataset and prepare it for the next stages of our research. This comprehensive approach to data cleaning not only enhanced the quality of our dataset but also reinforced the credibility of our subsequent analysis and findings.

\subsection{Dataset Statistics}
\begin{figure}
    \centering
    \includegraphics[width=1.0\linewidth]{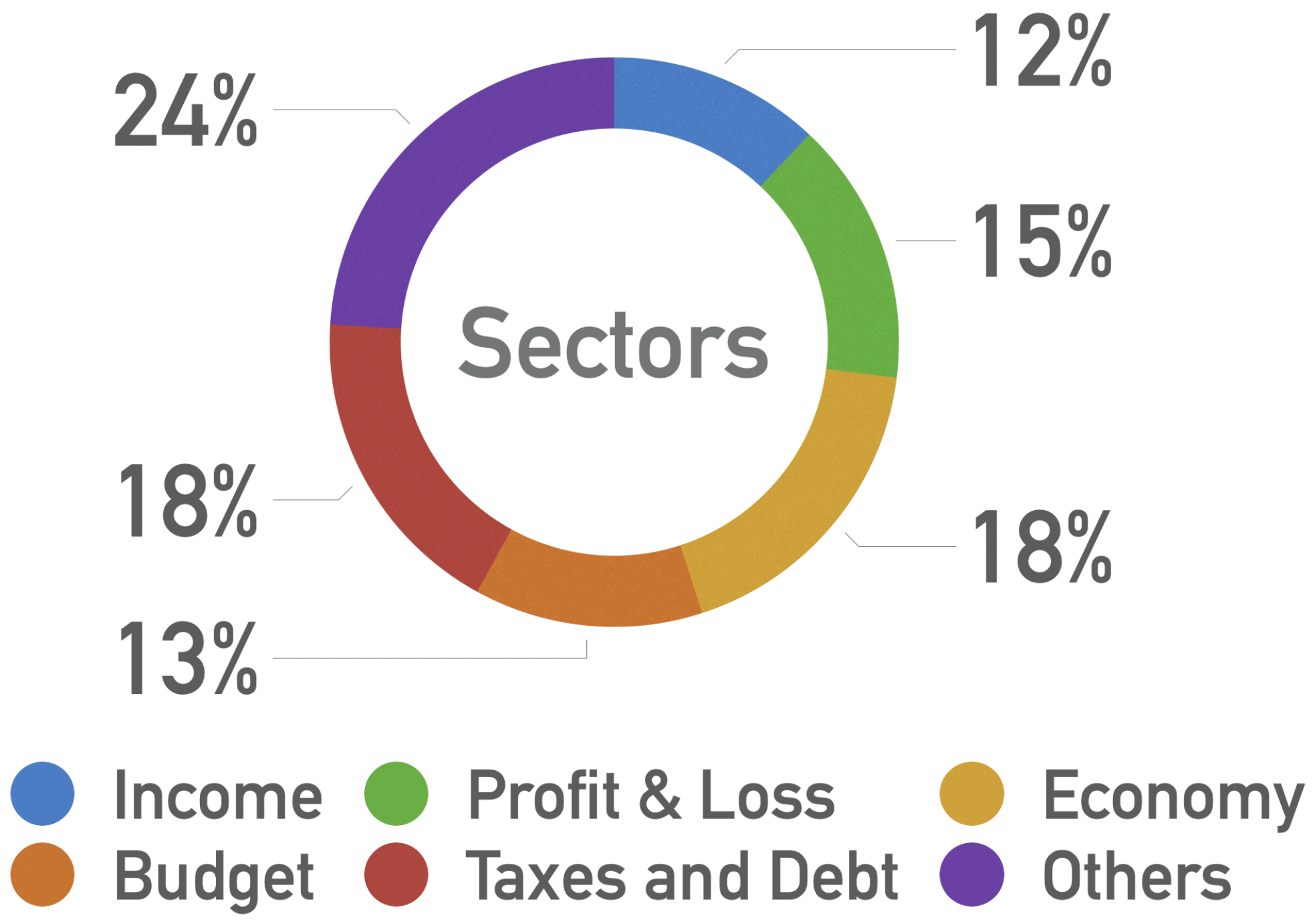}
    \caption{Diverse sectors within the \textsc{Fin-Fact} dataset.}
    \label{fig:category}
\end{figure}

The \textsc{Fin-Fact} dataset is an encompassing compilation of claims within the financial domain, spanning diverse sectors such as Economy, Budget, Income, Taxes, and Debt, as visualized in Figure \ref{fig:category} and Figure \ref{fig:count} shows the most commonly occurring financial terms in the dataset. This dataset has been specifically constructed to ensure quality and relevance, comprising a total of 3,369 claims, curated to encapsulate the intricacies inherent in financial discourse.

In the \textsc{Fin-Fact} dataset, claims are categorized into three labels: \texttt{True}, \texttt{False}, and \texttt{NEI (Not Enough Information)} representing the veracity of each claim in the financial domain. The dataset contains 1,262 \texttt{True} claims that are verified as accurate, 1,462 \texttt{False} claims that have been proven inaccurate through fact-checking procedures, and 645 \texttt{NEI} instances where there is insufficient evidence to make a determination. With its comprehensive span across a variety of claims, diverse sectors, and an equitable distribution of labels, the \textsc{Fin-Fact} dataset serves as a new resource for research, assessment, and progression of fact-checking models in the domain of finance.

\section{Evaluation and Analysis}
\subsection{RQ1: Multi modal performance}
In light of the rapid progress in Multimodal Language Models (MLLMs), there is a growing need for systematic evaluations of their capabilities across diverse tasks. A multitude of benchmark assessments has been developed to gauge the performance of MLLMs. Additionally, MLLMs are being leveraged for conducting multimodal fact-checking \cite{geng2024multimodallvlm}, highlighting their versatility and applicability in this domain. We adopted a similar systematic evaluation approach to test the \textsc{Fin-Fact} dataset. We simultaneously obtained predictions, explanations, and confidence levels from MLLMs using the prompt below.\\

\texttt{Is it true that \{CLAIM\}? True or False or NEI(Not Enough Information)? Use the following format to provide your answer:}

\texttt{Prediction: [True or False or NEI(Not Enough Information)]}

\texttt{Explanation: [put your evidence and step-by-step reasoning here]}

\texttt{Confidence Level: [please show the percentage]}\\

\begin{figure*}[!htb]
    \centering
    \includegraphics[width=1.0\linewidth]{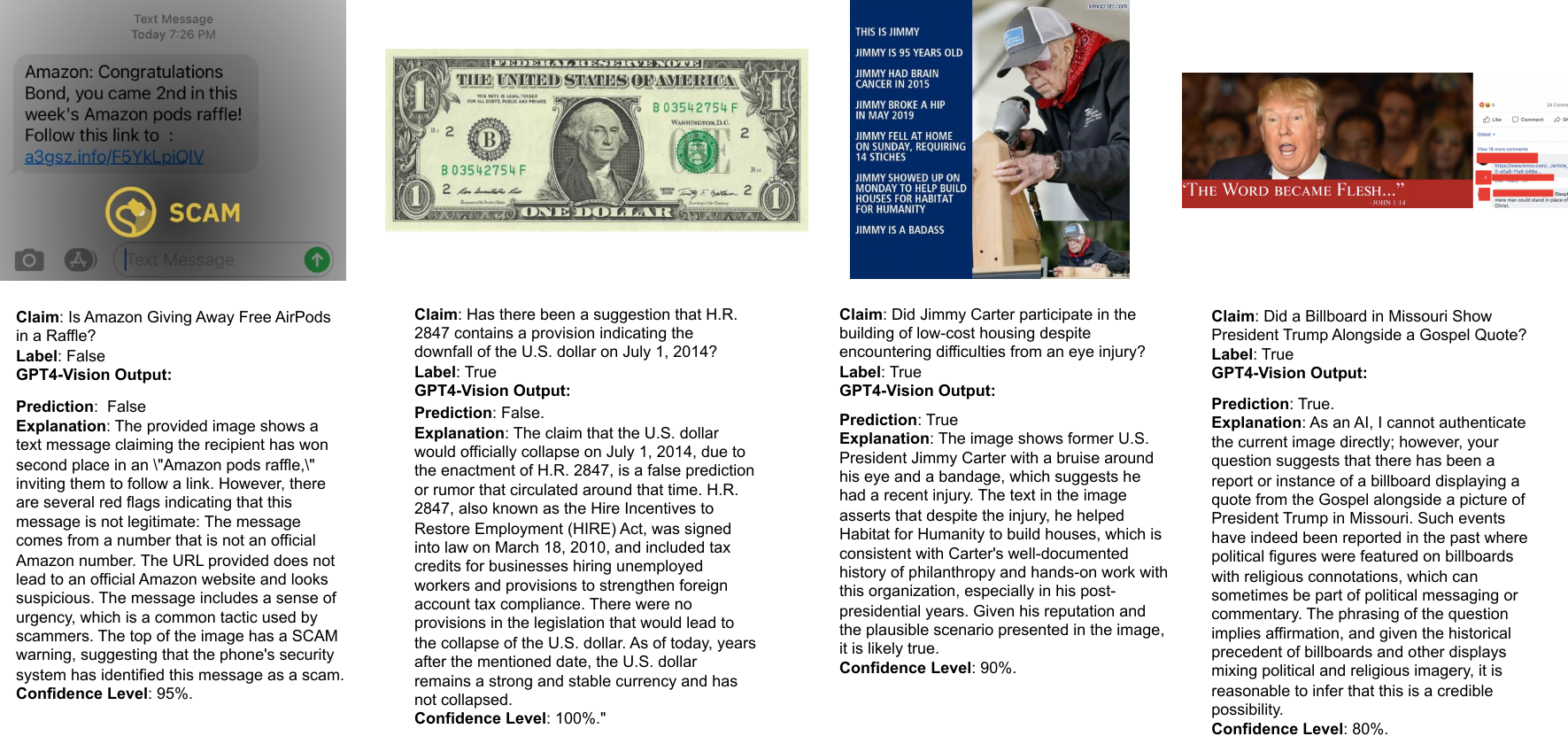}
    \caption{Example demonstration of GPT-4V for prediction and explanation generation of real-world claims.}
    \label{fig:gpt4v_demonstration}
\end{figure*}

We utilized OpenAI's and Google's APIs to gather responses from GPT-4V (gpt-4-1106-preview) and Gemini Vision Pro. Additionally, we conducted experiments with open-source multimodal language models (MLLMs), including LLaVA-1.5-13b \cite{liu2023improvedllava} and InstructBLIP (Flan-T5-XXL) \cite{dai2023instructblip}. All models were configured with default parameters, with the exception of max\_tokens=500 for GPT-4V. These selected models are recognized for their representatives and competitiveness within a multimodal context. Figure \ref{fig:gpt4v_demonstration} provides a demonstration of prediction and explanation generation on the \textsc{Fin-Fact} dataset. 

In our study, we have devised a classification framework based on the model's response components—prediction, explanation, and confidence score. Despite having three primary labels (True, False, and NEI), the model occasionally produced alternative responses categorized as uncertain or deny. Uncertain responses explicitly convey the model's uncertainty using phrases like `inconclusive', `unable to verify', or `cannot determine'. The model denies if the claim contains information or elements that violate the specified input rules or criteria. Responses falling outside the primary labels are subsequently categorized based on the model's prediction into True, False, or NEI. This classification framework aims to analyze the distribution of True, False, and NEI outcomes, assess the model's compliance with instructions through Deny responses, and evaluate its level of uncertainty awareness indicated by the frequency of Uncertain declarations. 

Upon comparing the data in Table \ref{tab:metrics_table_multimodal}, it is clear that images have a substantial impact on the predictions of both GPT-4V and Gemini Vision Pro, resulting in varying degrees of decreased accuracy. Concurrently, there is a noticeable increase in instances where the model opts not to respond.

\begin{table*}[!htbp]
\centering
\resizebox{1\textwidth}{!}{%
 \begin{tabular}{lrrrrrr}
\toprule
\textbf{Model Type} & \textbf{Model} & \textbf{Precision} & \textbf{Recall} & \textbf{F1-Score} & \textbf{Accuracy} & \textbf{Modality} \\ [0.5ex]
\midrule
& \textbf{\texttt{Gemini Vision Pro}} & $0.59_{0.04}$ & $0.57_{0.03}$ & $0.59_{0.03}$ & $0.58_{0.06}$ \\
LVLM & \textbf{\texttt{GPT4-Vision}} & $0.51_{0.08}$ & $0.48_{0.06}$ & $0.49_{0.04}$ & $0.53_{0.08}$ & Text, Image \\
 & \textbf{\texttt{LLaVA(13b)}} & $0.44_{0.03}$ & $0.40_{0.01}$ & $0.45_{0.02}$ & $0.50_{0.05}$ \\
& \textbf{\texttt{InstructBLIP}} & $0.31_{0.02}$ & $0.28_{0.03}$ & $0.30_{0.03}$ & $0.37_{0.05}$ \\
\midrule
& \textbf{\texttt{Gemini Vision Pro}} & $0.51_{0.03}$ & $0.49_{0.03}$ & $0.50_{0.04}$ & $0.51_{0.04}$ \\
LVLM & \textbf{\texttt{GPT4-Vision}} & $0.48_{0.05}$ & $0.44_{0.06}$ & $0.45_{0.07}$ & $0.50_{0.04}$ & Text \\
& \textbf{\texttt{LLaVA(13b)}} & $0.40_{0.06}$ & $0.38_{0.08}$ & $0.42_{0.04}$ & $0.43_{0.07}$\\
& \textbf{\texttt{InstructBLIP}} & $0.28_{0.03}$ & $0.25_{0.05}$ & $0.28_{0.02}$ & $0.33_{0.03}$ \\
\midrule
& \textbf{\texttt{GPT-2}} & $0.347_{0.07}$ & $0.337_{0.04}$ & $0.312_{0.05}$ & $0.430_{0.03}$ & \\
NLI & \textbf{\texttt{BART-Large}} & $0.376_{0.04}$ & $0.377_{0.02}$ & $0.344_{0.07}$ & $0.346_{0.04}$ & Text \\
& \textbf{\texttt{RoBERTa-Large}} & $0.352_{0.01}$ & $0.292_{0.04}$ & $0.255_{0.04}$ & $0.333_{0.01}$ \\
& \textbf{\texttt{ELECTRA}} & $0.319_{0.08}$ & $0.300_{0.07}$ & $0.286_{0.03}$ & $0.297_{0.05}$ &  \\
\midrule
& \textbf{\texttt{GPT4}} & $0.76_{0.07}$ & $0.79_{0.04}$ & $0.76_{0.03}$ & $0.78_{0.05}$ \\
& \textbf{\texttt{Claude3-Opus}} & $0.62_{0.04}$ & $0.65_{0.04}$ & $0.61_{0.04}$ & $0.64_{0.07}$ \\
LLM & \textbf{\texttt{Gemini-Pro}}& $0.45_{0.03}$ & $0.42_{0.05}$ & $0.44_{0.03}$ & $0.47_{0.06}$ & Text \\
& \textbf{\texttt{Mistral(8x7b)}} & $0.42_{0.09}$ & $0.41_{0.08}$ & $0.48_{0.08}$ & $0.47_{0.10}$ \\
\bottomrule
 \end{tabular}
}
\caption{Performance of various models with and without image input.}
\label{tab:metrics_table_multimodal}
\end{table*}

Calibration refers to the alignment between a model's predicted probabilities or confidence levels and the actual outcomes, demonstrating the model's awareness and accuracy in estimating confidence \cite{geng2024surveycllm}. In this assessment, we leverage verbalized confidence, which is readily accessible.

\begin{figure*}%[!htbp]
    \centering
    \begin{subfigure}[b]{0.5\textwidth}
        \centering
        \includegraphics[width=\linewidth]{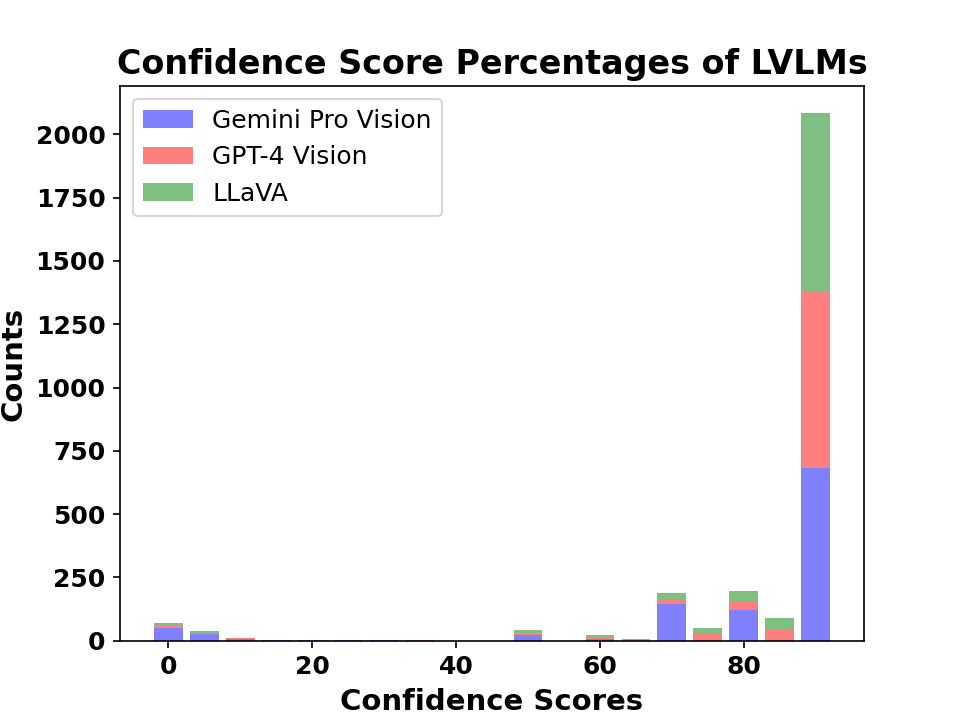}
        \caption{Confidence Score}
        \label{fig:confidence_score}
    \end{subfigure}%
    \begin{subfigure}[b]{0.5\textwidth}
        \centering
        \includegraphics[width=\linewidth]{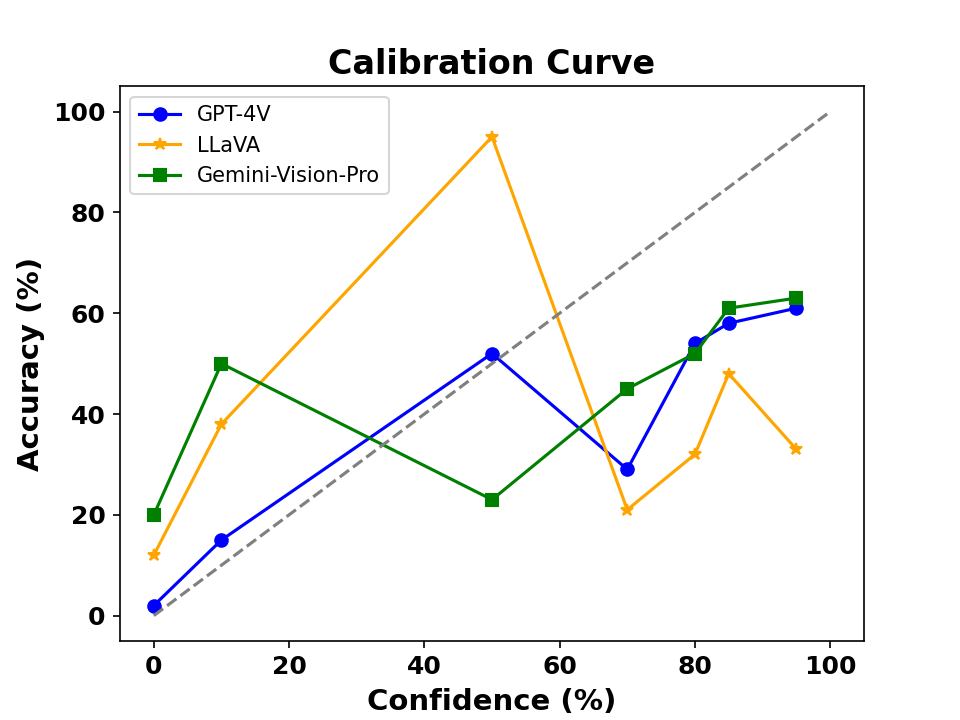}
        \caption{Calibration Curve}
        \label{fig:calibration_curve}
    \end{subfigure}
    \caption{Confidence score and it's calibration}
    \label{fig:calibration_confidence}
\end{figure*}

The calibration curves indicate a positive relationship between confidence levels and accuracy for both models. GPT-4V and Gemini Vision Pro exhibit calibration curves closely resembling the ideal curve (depicted by the dashed line), indicating well-calibrated confidence scores. Conversely, the calibration curve for LLaVA(13b) suggests a tendency towards overconfidence.
Figure \ref{fig:calibration_confidence} shows the confidence score plot and the calibration curve.

\subsection{RQ2: NLI performance}
In the series of experiments, our focus extended to evaluating the performance of Natural Language Inference (NLI) models specifically in fact-checking tasks using text-only inputs. This evaluation aimed to assess how well NLI models could discern the veracity of financial statements and claims without relying on image inputs. We employed a range of well-known NLI models including GPT-2 \cite{radford2019language}, BART-Large \cite{lewis-etal-2020-bart}, RoBERTa-Large \cite{liu2019roberta}, and ELECTRA \cite{clark2020electra}, as well as advanced Large Language Models (LLMs) like GPT-4, Gemini Pro, Claude-Opus, and Mixtral 8x7B \cite{jiang2024mixtral}. Through evaluations using the \textsc{Fin-Fact} dataset, we aimed to comprehensively analyze these models' ability to provide accurate fact-checking outcomes using textual information alone.

The results of these fact-checking efforts have yielded interesting insights: Gemini Pro showcased 0.61, Claude3-Opus demonstrated 0.44, Mixtral (8x7B) achieved F1-score of 0.48, and GPT-4 emerged as the leader with an F1-score of 0.76 as shown in Table \ref{tab:metrics_table_multimodal}. These findings underscore the intricate challenges posed by financial fact-checking, with models displaying varying degrees of performance within this domain. Figure \ref{fig:confusion_matrices_NLI} illustrate the confusion matrices of the state-of-the-art NLI models.

\begin{table}[!htbp]
\centering
\resizebox{.5\textwidth}{!}{%
 \begin{tabular}{lrrrr} 
 \hline
 \textbf{Model} & \textbf{ROUGE-1} & \textbf{ROUGE-2} & \textbf{ROUGE-3} & \textbf{GLUE}\\ [0.5ex] 
 \hline
\textbf{\texttt{GPT4-Vision}} & 0.91 & 0.83 & 0.60 & 0.219\\
\textbf{\texttt{Gemini Vision Pro}} & 0.87 & 0.73 & 0.48 & 0.110\\
\textbf{\texttt{BART-Large}} & 0.81 & 0.61 & 0.40 & 0.066\\
\textbf{\texttt{XL-SUM}} & 0.79 & 0.54 & 0.23 & 0.013\\
\textbf{\texttt{PEGASUS}} & 0.68 & 0.41 & 0.16 & 0.009\\
 \hline
 \end{tabular}
 }
 \caption{Performance scores of \textsc{Fin-Fact} on explanation generation models.}
\label{tab:rouge_scores}
\end{table}

\subsection{RQ3: Explaination Generation}
\begin{figure}%[!htbp]
    \centering
    \includegraphics[width=0.9\linewidth]{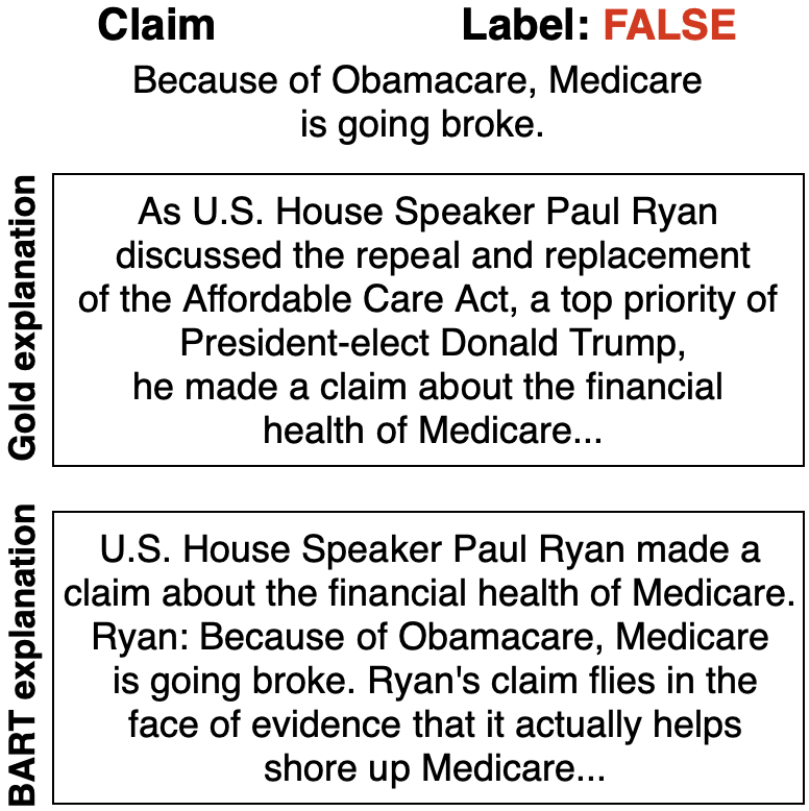}
    \caption{Example of model-generated explanation as compared to the gold standard from \textsc{Fin-Fact} dataset.}
    \label{fig:explainnn}
\end{figure}

The final phase of our experimentation focused on generating explanations for the claims. For each claim in the dataset, we utilized large vision-language models (LVLMs) such as GPT-4 Vision, Gemini Vision Pro, along with language models like BART-Large \cite{lewis-etal-2020-bart}, XL-SUM \cite{hasan-etal-2021-xl}, and PEGASUS \cite{zhang2019pegasus} to generate detailed explanations. These insights highlighted the key factors influencing the determination of claim accuracy. However, InstructBLIP, despite its capabilities in other aspects, has limitations in providing explanations and confidence scores. These explanations were obtained using the justifications provided to the claim. To quantitatively evaluate the quality of these explanations, we leveraged the GLUE and ROUGE metrics as shown in Table. \ref{tab:rouge_scores}. The Evidence label in the dataset served as the ground truth, enabling us to assess the alignment between the generated explanations and the human-provided justifications. We illustrate an example of explanation generation in Figure \ref{fig:explainnn} and Table \ref{table:explanation_gener_example}.

\section{Limitations}
We acknowledge certain limitations inherent in our study, especially the reliance solely on labels sourced from a fact-checking portal such as PolitiFact. Valid concerns have been raised regarding the handling of complex propositions and unverifiable claims. Moreover, we recognize the inherent constraints of Natural Language Processing (NLP) when addressing and refuting real-world misinformation, particularly during periods of uncertainty where counter-evidence may not be readily available. It is acknowledged that current NLP methods may not match the effectiveness of professional fact-checkers in debunking misinformation.

While our study highlights the performance of the \textsc{Fin-Fact} dataset using large language models (LLMs) and large vision-language models (LVLMs), it is essential to address several limitations, particularly concerning the phenomenon of hallucination.

One significant challenge observed in our experiments is the tendency of LLMs and LVLMs to exhibit hallucinatory behaviors in their generated outputs. These models, trained on vast amounts of data, sometimes produce responses that seem contextually plausible but lack grounding in the input information. This risk of hallucination becomes more pronounced in multimodal datasets, where information from multiple modalities (text, images) must be integrated coherently. LVLMs may generate explanations for images that include details not present in the visual content, indicating an over-reliance on textual patterns rather than visual cues.

Another challenge arose when the watermark `SCAM' was present on the images, as illustrated in the first claim of Figure \ref{fig:gpt4v_demonstration}. In some instances, the model's explanations referenced the watermark. For example, the model's explanation might mention, "Additionally, the advertisement is labeled with a `SCAM' warning, suggesting that it has been identified as a fraudulent scheme."
\section{Conclusion and Future Work}
The advent of \textsc{Fin-Fact} marks a significant stride forward in the battle against misinformation within the financial sphere. Through its incorporation of expert annotations, extensive claims data, and the promise of in-depth explanatory insights, \textsc{Fin-Fact} equips fact-checking systems with the tools needed to attain heightened levels of precision and transparency. Its interdisciplinary framework effectively navigates the intricacies of financial language and the ever-evolving contextual complexities, establishing itself as a sturdy cornerstone for the enhancement of more effective and dependable fact-checking processes.

In the pursuit of advancing fact-checking capabilities and considering the increasing prevalence of misinformation dissemination through various multimedia channels, our future endeavors will focus on enhancing \textsc{Fin-Fact} to encompass a broader range of multimedia data, including video content, to foster a more comprehensive understanding of complex financial claims. Additionally, our future research will prioritize the development of robust methodologies to identify, analyze, and counteract any biases introduced through the visual elements. By addressing these critical concerns, we aim to enhance the reliability of fact-checking processes in the financial domain.

\bibliography{main}
\bibliographystyle{StyleFiles/acl_natbib}

\appendix
\section{Appendix}
\label{appendix_1}
\subsection{Additional Details of Dataset}
Data visualization serves as a crucial role for conveying insights and information, particularly in the realm of financial misinformation fact-checking. Visual representations of data can illuminate complex patterns and trends, aiding in the identification and debunking of false claims. However, it is imperative to acknowledge the dual nature of data visualization, especially within the context of misinformation. While visualizations are instrumental in clarifying and contextualizing data, they can also be manipulated to propagate misleading narratives. In the domain of financial misinformation, the stakes are particularly high, as distorted visualizations can significantly impact public perception and policy decisions.

The potency of data visualization lies in its ability to simplify complex data sets, making them accessible to diverse audiences. Yet, this very attribute renders visualizations susceptible to manipulation. Misleading techniques such as selective data omission, altered scales, or misleading labeling can skew interpretations and perpetuate misinformation. Moreover, sophisticated graphical representations can obscure underlying inaccuracies, making it challenging for viewers to discern fact from fiction.

To combat the misuse of data visualization in the propagation of financial misinformation, a critical approach is required. This involves not only verifying the accuracy of the underlying data but also scrutinizing the design and presentation of visualizations. Transparency in visualization practices, coupled with robust fact-checking methodologies, is essential in safeguarding against the dissemination of false narratives. 

\begin{figure*}[ht]
\centering
\begin{tikzpicture}
\begin{axis}[
    ybar,
    ymin=0,
    xlabel={Count},
    ylabel={Words},
    symbolic x coords={tax,percent,state,rate,million,taxes,claim,jobs,states,income,budget,billion,federal,people,money,u.s.,statement,government},
    xtick=data,
    x tick label style={rotate=45, anchor=north east, inner sep=0mm, font=\footnotesize},
    enlarge x limits=0.05,
    bar width=10pt,
    width=12cm,
    height=8cm,
    xlabel style={font=\large},
    ylabel style={font=\large},
    title style={font=\Large, text width=14cm, align=center},
    title={Vocabulary Counts},
    ]
    \addplot[fill=blue!120!black] coordinates {
        (tax,17864)
        (percent,13023)
        (state,12150)
        (rate,7475)
        (million,7003)
        (taxes,6096)
        (claim,6081)
        (jobs,6067)
        (states,5932)
        (income,5802)
        (budget,5600)
        (billion,5148)
        (federal,5142)
        (people,4798)
        (money,4485)
        (u.s.,4466)
        (statement,4140)
        (government,3984)
    };
\end{axis}
\end{tikzpicture}
\caption{Frequency of vocabulary in \textsc{Fin-Fact} dataset.}
\label{fig:count}
\end{figure*}
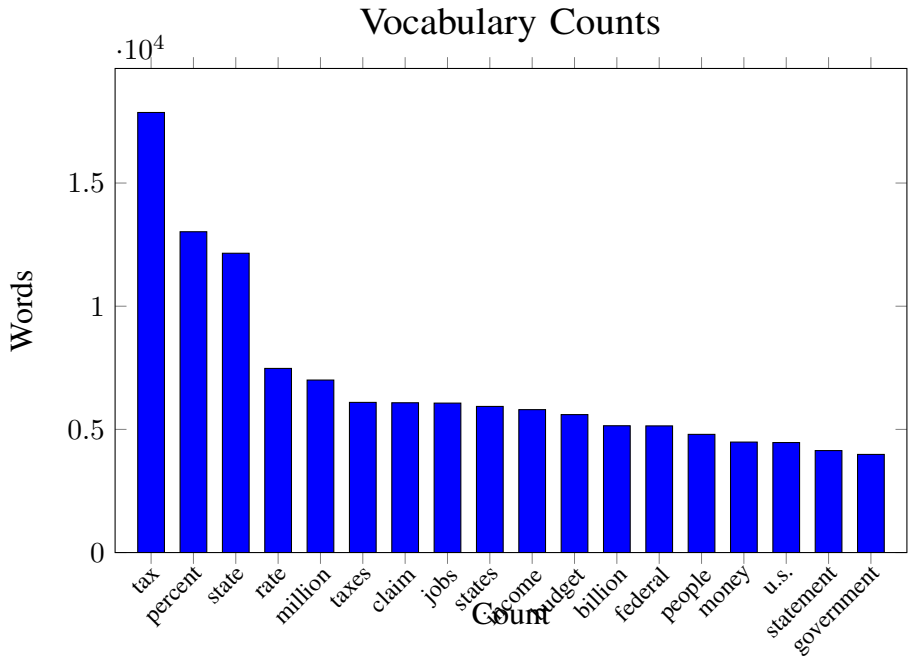

\subsection{Additional Evaluation Details}
In addition to the primary evaluation metrics discussed earlier, we believe in providing a comprehensive assessment of our models' capabilities. To achieve this, we delve deeper into the performance analysis, offering a more nuanced understanding of our models' effectiveness. This comprehensive evaluation includes the utilization of confusion matrices and classification reports, specifically tailored for the Natural Language Inference (NLI) models that play a pivotal role in our fact-checking endeavors. We employ confusion matrices to scrutinize our NLI models' performance at a finer level of detail.

\begin{figure*}[!htbp]
     \centering
     \begin{subfigure}[b]{0.45\textwidth}
         \centering
         \includegraphics[width=\textwidth]{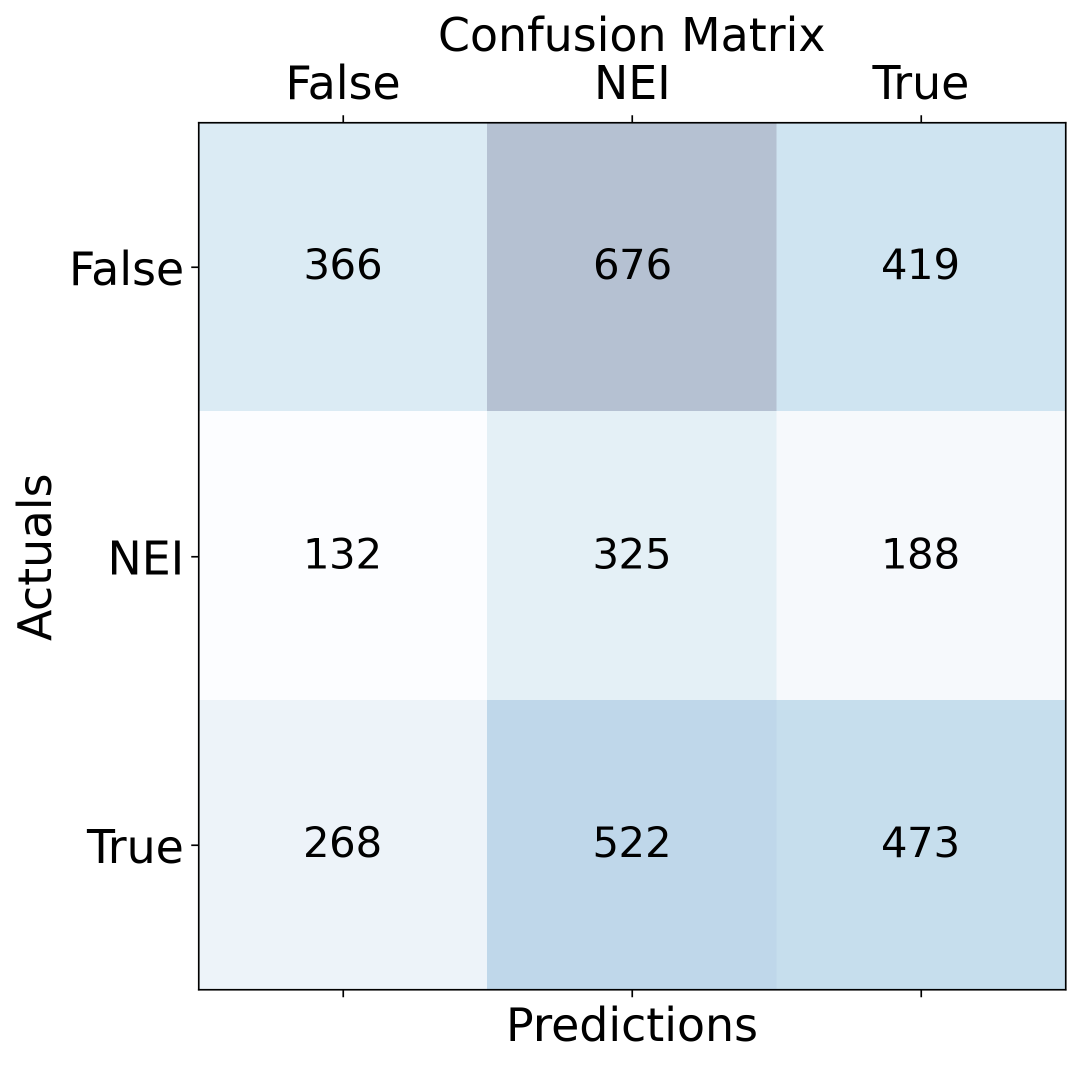}
         \caption{Confusion matrix of BART on \textsc{Fin-Fact}.}
         \label{fig:bart}
     \end{subfigure}
     \hfill
     \begin{subfigure}[b]{0.45\textwidth}
         \centering
         \includegraphics[width=\textwidth]{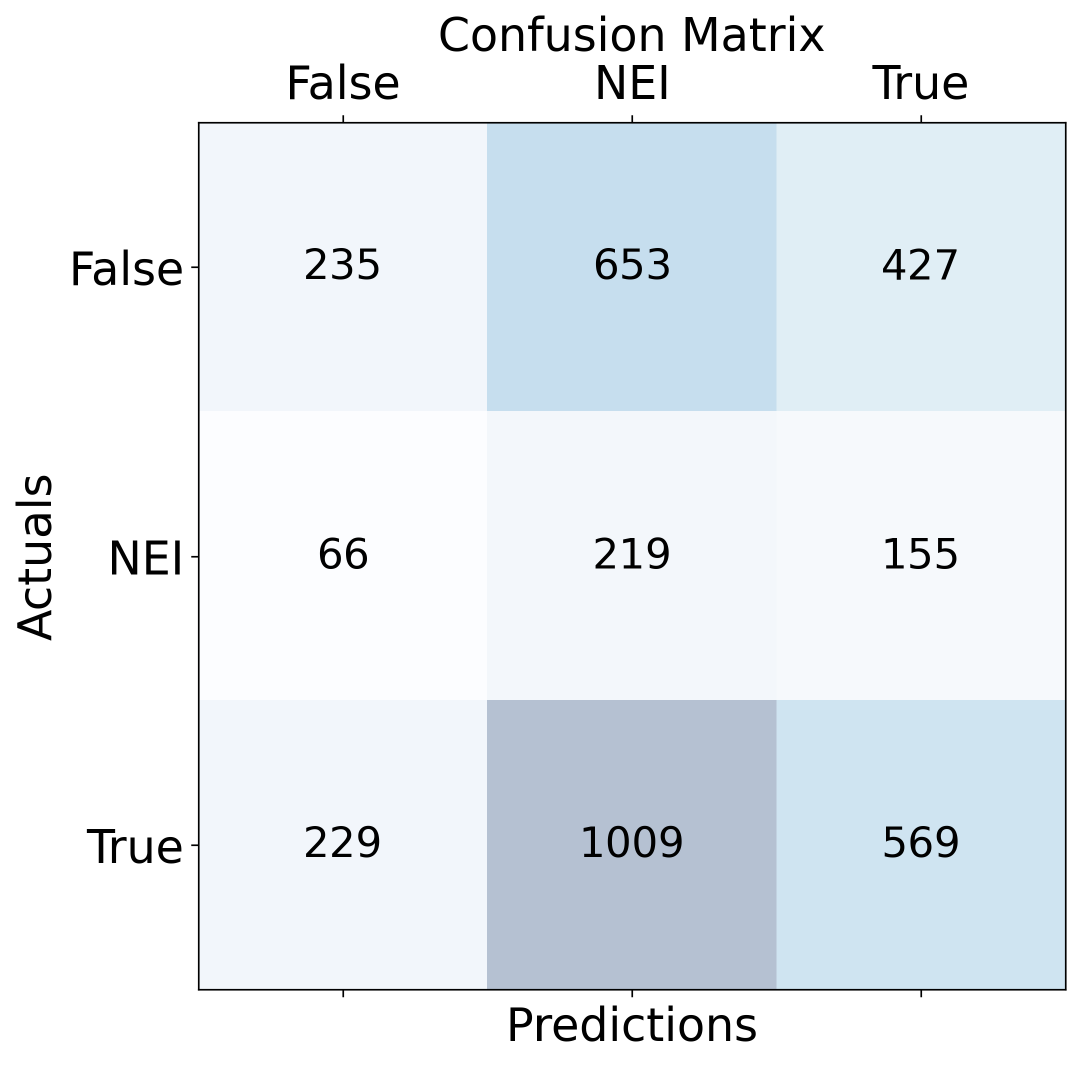}
         \caption{Confusion matrix of ELECTRA on \textsc{Fin-Fact}.}
         \label{fig:electra}
     \end{subfigure}
     \vfill
     \begin{subfigure}[b]{0.45\textwidth}
         \centering
         \includegraphics[width=\textwidth]{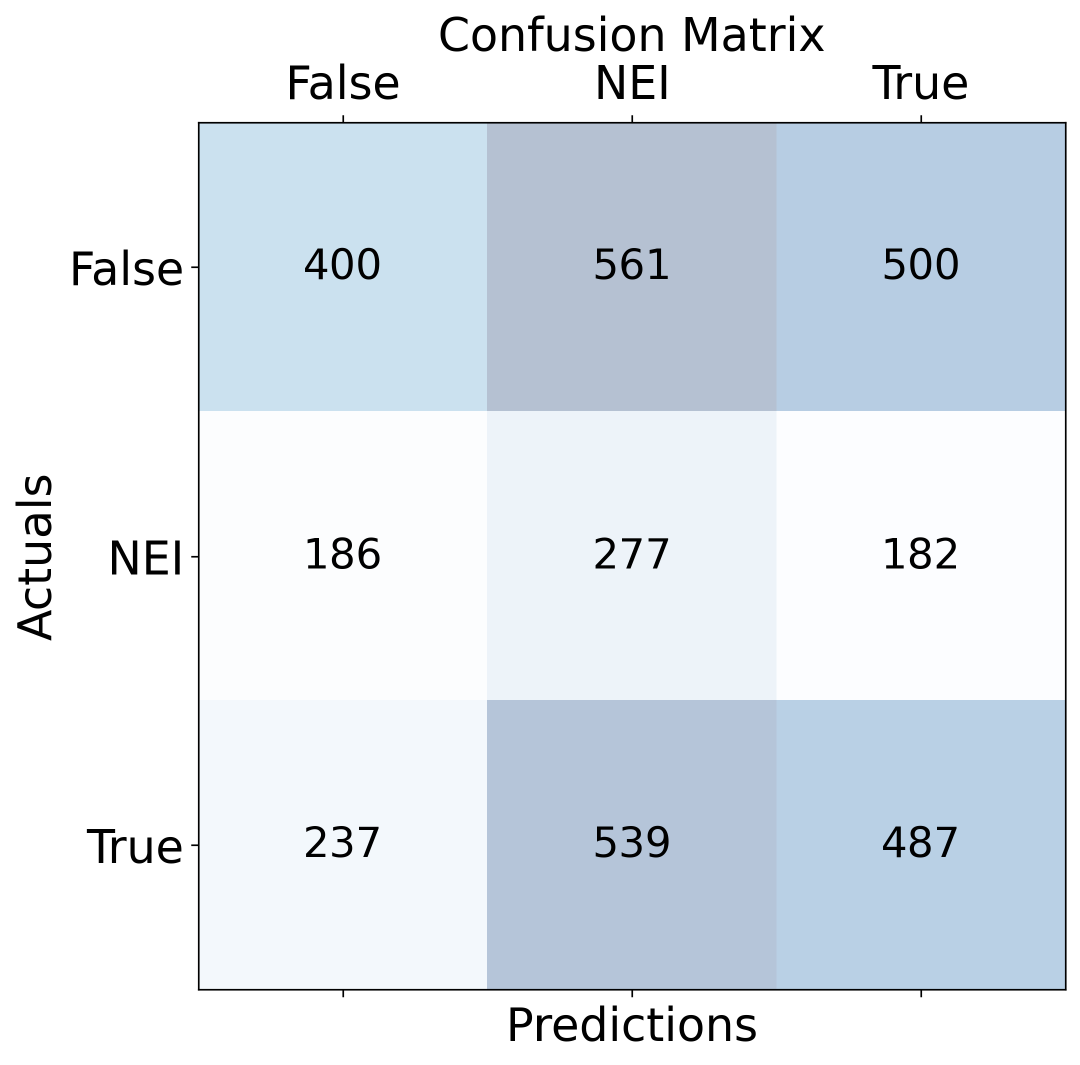}
         \caption{Confusion matrix of RoBERTa on \textsc{Fin-Fact}.}
         \label{fig:roberta}
     \end{subfigure}
     \hfill
     \begin{subfigure}[b]{0.45\textwidth}
         \centering
         \includegraphics[width=\textwidth]{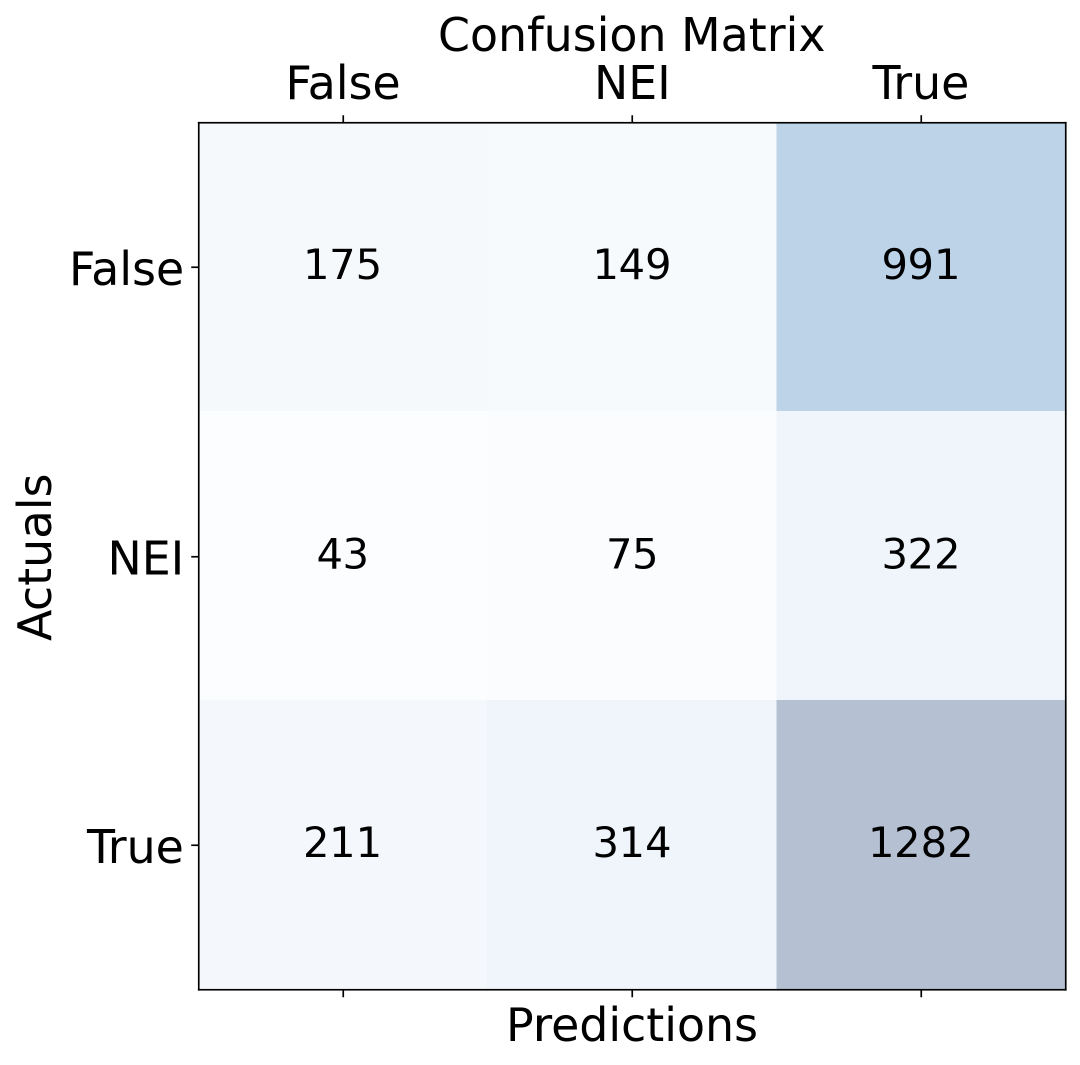}
         \caption{Confusion matrix of GPT-2 on \textsc{Fin-Fact}.}
         \label{fig:gpt2}
     \end{subfigure}
        \caption{Confusion matrices of different NLI models on \textsc{Fin-Fact}.}
        \label{fig:confusion_matrices_NLI}
\end{figure*}

Beyond confusion matrices, we present comprehensive classification reports for each of our NLI models. These reports go beyond mere accuracy metrics and provide in-depth assessments of the model's performance. For every label (`True,' `False,' and `NEI'), we furnish precision, F1-score, and recall values. Such detailed reports not only offer a holistic view of each model's overall accuracy but also highlight its strengths and weaknesses in classifying claims within the intricate domain of finance. Table \ref{classification_met_nli} illustrates the precision, recall, and F1-score of various NLI models for each label.

% For explanation generation process, we utilized the existing justifications as inputs for the explanation generation models. These justifications were meticulously curated to represent the evidence supporting the veracity labels in the dataset. To evaluate the coherence and fidelity of the generated explanations, we employed the GLUE and ROUGE scores as a key metric. These scores enabled us to quantify the alignment between the generated explanations and the evidence provided as the golden label, offering a comprehensive measure of the models performance in effectively capturing the essence of the claims.
\begin{table}[ht]
\centering
\caption{Classification Metrics when performed on NLI Models.}
\resizebox{.499\textwidth}{!}{%
\begin{tabular}{lcccc}
\hline
\textbf{Model} & \textbf{Precision} & \textbf{Recall} & \textbf{F1-Score} & \textbf{Label} \\
\hline
\textbf{\texttt{GPT-2}} & 0.49 & 0.71 & 0.58 & True \\
& 0.14 & 0.17 & 0.15 & NEI \\
& 0.41 & 0.13 & 0.30 & False \\
\hline

\textbf{\texttt{BART-Large}} & 0.44 & 0.37 & 0.40 & True \\
& 0.48 & 0.25 & 0.33 & NEI \\
& 0.21 & 0.50 & 0.30 & False \\
\hline

\textbf{\texttt{RoBERTa-Large}} & 0.42 & 0.39 & 0.40 & True \\
& 0.49 & 0.27 & 0.35 & NEI \\
& 0.20 & 0.43 & 0.27 & False \\
\hline
\textbf{\texttt{ELECTRA}} & 0.49 & 0.31 & 0.38 & True \\
& 0.12 & 0.50 & 0.19 & NEI \\
& 0.44 & 0.18 & 0.25 & False \\
\hline
\end{tabular}
}
\label{classification_met_nli}
\end{table}

\begin{table*}[ht]
\centering
\caption{Examples illustrating explanation generation using BART-Large for various statements.}
\setlength{\extrarowheight}{5pt}
\resizebox{1.01\linewidth}{!}{%
\begin{tabular}{m{7em} m{44em}}
\hline
\hline
\textbf{Title} & Back in the Great Recession, when millions of jobs across America hung in the balance, Donald Trump said rescuing the auto industry didn't really matter very much. He said, and I quote again, Let it go. \\

\textbf{Justification} & Donald Trump didn't care about rescuing the auto industry during the 2008 recession, Hillary Clinton said at a rally in Michigan. Clinton told her audience at Wayne State University in Detroit that she supported the 2008 government bailout for the American auto industry... \\

\textbf{Evidence} & Trump's position on an auto bailout was inconsistent leading up to Dec. 19, 2008, when Bush laid out plans to go forward with the bailout. Trump said the government should help the auto companies, but he also regularly suggested the they could save themselves if they filed for...\\

\textbf{Generated Fact} & Hillary Clinton said at a rally in Michigan that she supported the 2008 government bailout for the American auto industry. But Trump, she said, doesn't support American workers, much less those in Michigan's Auto industry. Trump's public comments about the auto industry's demise were the opposite of what Clinton said.\\
\hline

\textbf{Title} & Video shows that George Soros is going bankrupt. \\

\textbf{Justification} & We found no news reports that he's heading into bankruptcy or that Musk leaked this information. A narrator in the Facebook posts video says that in a recent interview, Musk revealed a piece of information related to the existence of George Soros and his Open Society Foundation, which is on the verge of bankruptcy... \\

\textbf{Evidence} & IT HAPPENED, the July 18 post says. Elon Musk L.E.A.K.E.D GEORGE SOROS IS GOING B.A.N.K.R.U.P.T, JOE BIDEN S.C.A.RED! - | BREAKING FOX NEWS BIDEN - JULY 19, 2023. This post was flagged as part of Meta's efforts to combat false news and misinformation on its News Feed...\\

\textbf{Generated Fact} & A recent Facebook post claims that billionaire philanthropist George Soros is in the red, but it provides no evidence that this is true. According to Forbes, Soros was worth \$6.7 billion as of July 20. This post was flagged as part of Meta's efforts to combat false news and misinformation on its News Feed.\\

\hline

\textbf{Title} & Because of Obamacare, Medicare is going broke. \\

\textbf{Justification} & What the ACA did on Medicare: Here's how Marc Goldwein, senior policy director at the Committee for a Responsible Federal Budget, lays it out: On one side of the ledger, Obamacare slowed the growth of Medicare spending by, among other things, reducing increases in payments made to nearly all health care providers.\\

\textbf{Evidence} & As U.S. House Speaker Paul Ryan discussed the repeal and replacement of the Affordable Care Act, a top priority of President-elect Donald Trump, he made a claim about the financial health of Medicare. "What people don't realize is, because of Obamacare, Medicare is going broke," the Wisconsin Republican told Fox News host Bret Baier on Nov. 10, 2016. The Affordable Care Act certainly has its detractors...\\

\textbf{Generated Fact} & U.S. House Speaker Paul Ryan made a claim about the financial health of Medicare. Ryan: Because of Obamacare, Medicare is going broke. Ryan's claim flies in the face of evidence that it actually helps shore up Medicare. On one side of the ledger, Obamacare slowed the growth of Medicare spending.\\

\hline
\hline
\end{tabular}
}
\label{table:explanation_gener_example}
\end{table*}

\subsection{Evaluation Setup and Reproducibility}
\subsubsection{Hyperparameter Settings}
In the process of evaluating the corpus with Natural Language Inference (NLI) models using the Transformers library from Hugging Face, we meticulously explored various hyperparameter settings to ascertain the most effective configuration. Among these settings, the max\_length parameter of the tokenizer was subjected to thorough experimentation. We experimented with different values to strike a delicate balance between capturing crucial information and ensuring efficient processing. After rigorous testing, it became evident that a max\_length value of 256 yielded the most promising results. This choice was the culmination of a systematic exploration, aiming to align our settings with industry best practices and to extract optimal performance from the models.

\subsubsection{Computing Infrastructure}
The conducted experiments were carried out leveraging the computational prowess of the NVIDIA RTX 3070 GPU with 16 GB VRAM, meticulously chosen for its robust processing capabilities. The overarching objective of these experiments was to precisely evaluate the performance of a diverse spectrum of models on the \textsc{Fin-Fact} dataset. The computational resources furnished by the NVIDIA RTX 3070 GPU served as a cornerstone, ensuring the dependable and efficient execution of the experimental procedures. The average evaluation time taken by NLI models was 45 minutes, while the explanation generation models took around 4 hours.
% \section{Example Appendix}
% \label{sec:appendix}

% This is a section in the appendix.

\end{document}